\documentclass{article}

\usepackage{arxiv}

\usepackage{pgfplots}

\usepackage[utf8]{inputenc} 
\usepackage[T1]{fontenc}    
\usepackage{hyperref}       
\usepackage{url}            
\usepackage{booktabs}       
\usepackage{amsfonts}       
\usepackage{nicefrac}       
\usepackage{microtype}      
\usepackage{graphicx}
\usepackage{natbib}
\usepackage{doi}
\usepackage{import}
\usepackage{wrapfig, }

\title{Workpiece Image-based Tool Wear Classification in Blanking Processes Using Deep Convolutional Neural Networks}

\author{ \href{https://orcid.org/0000-0001-5743-8802}{\includegraphics[scale=0.06]{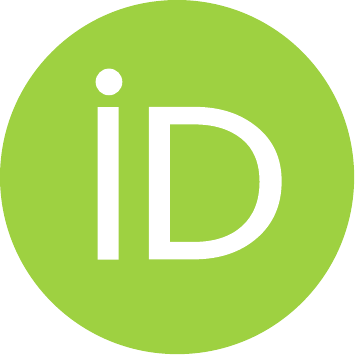}\hspace{1mm}Dirk Alexander Molitor}\\
	Institute for Production Engineering\\ and Forming Machines\\
	Technical University of Darmstadt\\
	Germany, 64289 Darmstadt\\
	\texttt{dirk.molitor@ptu.tu-darmstadt.de} \\
	\And
	\href{https://orcid.org/0000-0002-8695-714X}{\includegraphics[scale=0.06]{orcid.pdf}\hspace{1mm}Christian Kubik} \\
	Institute for Production Engineering\\ and Forming Machines\\
	Technical University of Darmstadt\\
	Germany, 64289 Darmstadt\\
	\texttt{kubik@ptu.tu-darmstadt.de} \\
	\And
	{\includegraphics[scale=0.06]{orcid.pdf}\hspace{1mm}Ruben Helmut Hetfleisch} \\
	Technical University of Darmstadt\\
	Germany, 64289 Darmstadt\\
	\texttt{rubenhelmut.hetfleisch@stud.tu-darmstadt.de} \\
	\And
	\href{https://orcid.org/0000-0001-7927-9523}{\includegraphics[scale=0.06]{orcid.pdf}\hspace{1mm}Peter Groche}\\
	Institute for Production Engineering\\ and Forming Machines\\
	Technical University of Darmstadt\\
	Germany, 64289 Darmstadt\\
	\texttt{groche@ptu.tu-darmstadt.de}\\
}

\renewcommand{\shorttitle}{\textit{arXiv} Template}

\hypersetup{
pdftitle={A template for the arxiv style},
pdfsubject={q-bio.NC, q-bio.QM},
pdfauthor={David S.~Hippocampus, Elias D.~Striatum},
pdfkeywords={Tool Condition Monitoring, Deep Learning, Wear Detection, Smart Manufacturing, Industry 4.0, Blanking},
}

\begin{document}
\maketitle
\setlength\intextsep{2pt}
\begin{abstract}
	Blanking processes belong to the most widely used manufacturing techniques due to their economic efficiency. Their economic viability depends to a large extent on the resulting product quality and the associated customer satisfaction as well as on possible downtimes. In particular, the occurrence of increased tool wear reduces the product quality and leads to downtimes, which is why considerable research has been carried out in recent years with regard to wear detection. While processes have widely been monitored based on force and acceleration signals, a new approach is pursued in this paper. Blanked workpieces manufactured by punches with 16 different wear states are photographed and then used as inputs for Deep Convolutional Neural Networks to classify wear states. The results show that wear states can be predicted with surprisingly high accuracy, opening up new possibilities and research opportunities for tool wear monitoring of blanking processes.
\end{abstract}


\section{Introduction}
Increasing quality requirements, high production rates and progressively more complex product geometries pose manufacturers with the challenges of a systematic automation and an efficient monitoring of blanking processes. Therefore, sensors are increasingly integrated into the processes and attempts are made to identify correlations between process anomalies and features of the recorded time series. Conventional approaches monitor the time series with the help of thresholds \citep{jemielniak2012tool,faura1997criterion,sari2017preliminary}), linear discriminant functions \citep{Lee.1997} or envelope curves and can thus distinguish binary process states from each other. In particular, the occurrence of punch wear, which has a negative impact on the resulting product quality \citep{hambli2002relationships,hambli2003numerical,mucha2010experimental,maiti2000assessment,Kubik.2021}, is a widely researched application scenario. Knowledge about the current wear can help manufacturers to reduce downtimes and flexibly adapt maintenance intervals to the punch wear state. Modern monitoring approaches predict, for example, the edge radius of the punch by means of multiple linear regressions, whereby features from force signals are used as input variables \citep{Hoppe.2019}. 
\\Some authors predict that artificial intelligence and machine learning will have enormous effects on blanking processes \citep{zheng2019state,Klingenberg.2008}, while the first applications have already been presented. For example, neural networks can be used to approximate the effects of fluctuating process parameters on geometric product properties, whereby training data can be obtained through FEM simulations \citep{stanke2017setup,al2012modeling,stanke2018predictive,hambli2002prediction,hambli2003application}. It was also shown that wear conditions can be accurately classified by extracting features from force signals and inputting them into Support Vector Machines \citep{kubik2021smart}.
\\Such time-series-based monitoring approaches have two different key disadvantages. On the one hand, the sensors are heavily loaded during the process due to high force and acceleration peaks, which can cause contact losses or slipping.  On the other hand, they need to be integrated as close as possible to the tool in order to realistically display the dynamic effects of the process \citep{Groche.2019}. To circumvent these issues, other data that contain correlations to wear conditions must be recorded and processed in real time.
\\ An efficient approach to overcome this problem is the recording of image data. Image-based sensors enable contactless monitoring of production processes and depict optically perceptible process characteristics in great detail \citep{cao2019manufacturing}. In combination with the advances made in the field of deep learning over the past decade, images can be processed in real time and correlations with labels recorded in production processes can be explored. However, to the best of the authors' knowledge, no image-based, data-driven monitoring approaches of blanking processes can be found in the literature. Therefore, this paper shows how images of blanked workpieces can be used to predict wear conditions of the punch. For this purpose, 7440 photos of 1860 workpieces, produced by punches with 16 different edge radii, are taken and used as inputs of Convolutional Neural Networks (CNN). Subsequently, it is demonstrated to what extent the CNN can classify the workpiece images according to the 16 wear states.
\begin{wrapfigure}[25]{o}{0.5\textwidth}
	\fontsize{9pt}{9pt}\selectfont
	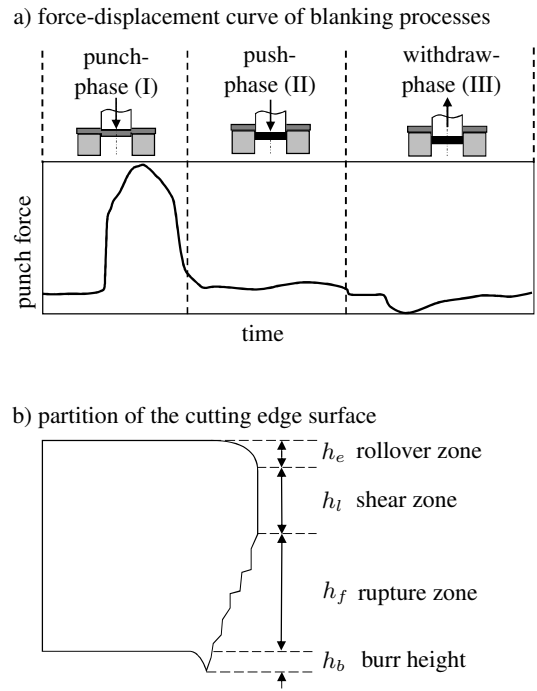
	\caption{Phases of the blanking process indicated by a force-displacement curve \citep{Hoppe.2019}  (a) and forming zones of the cutting edge surface \citep{VDI2906} (b)}
	\label{fig:blanking_process}
\end{wrapfigure}
\\The paper is organized as follows. Section 2 gives an overview of the blanking process, its wear phenomena, corresponding monitoring approaches and basics about CNNs. Section 3 presents the methodology consisting of the experimental setup, data generation process and CNN based modelling. Section 4 presents the results obtained from a pre-trained and a self-created CNN as well as their performance comparison. Finally, the results are summarised in section 5 and new research opportunities are discussed.

\section{Wear Monitoring in Blanking Processes}
\label{sec:1}
\subsection{Blanking processes}
Blanking is one of the most frequently used  sheet metal forming process and is part of the value chain in the automotive, information technology, power electronics and consumer goods sector. Blanking is a manufacturing process in which the final workpiece is separated from a sheet by applying a shearing force \citep{DIN_8588}. In the blanking process, the entire outer geometry of the workpiece is obtained in one working stroke whereby grid-shaped discard is left on the sheet metal strip \citep{lange1996handbook}. Although, blanking is a separation process according to DIN 8588, the literature classifies this manufacturing technique as a sheet metal forming process.

As shown in Figure \ref{fig:blanking_process} (a), blanking processes can be divided into three phases according to their force displacement curve \citep{Kubik.2021}. In the punch-phase (I), the tool hits the sheet and starts to elastically deform the system consisting of tool, material and press. If the stresses are further increased the material tends to a plastic deformation until the shearing stresses exceed the shear fracture limit, the material breaks and the elastic energy stored in the systems is abruptly released. In the push-phase (II), the workpiece is completely separated from the discard and pushed through the sheet metal strip. At the end of the push-phase the tool passes the bottom dead center and is pulled out of the die during the withdraw-phase (III). While the maximal forces during the punch-phase depend on process parameters (clearance, cutting edge radii, cutting line), material properties (tensile strength, sheet thickness) or press settings (stroke speed, stiffness of the press), the pushing as well as the withdrawing forces are mainly due to contact normal forces influenced by ongoing wear on the lateral punch surface \citep{Kubik.2021,Hohmann.2017}. 
As an indicator for the quality of blanked workpieces, the cutting edge surface is quantified according to Lange \citep{lange1996handbook}. As shown in Figure \ref{fig:blanking_process} (b), the form error found on the blanked surface is divided into rollover zone $h_e$, shear zone $h_l$ and rupture zone $h_f$. In addition, the burr height $h_b$ is a key indicator of poor product quality and is directly influenced by the wear state \citep{Feistle.2017}. 

\subsection{Wear phenomena during blanking}
Failures of blanking tools are often caused by major wear mechanisms referred to in the literature as adhesion, abrasion, surface fatigue and tribochemical wear \citep{lange1996handbook}. Especially, abrasive wear related to the rounding of the punch cutting edge due to the trend of processing high-strength materials is a major issue manufactures have to deal with \citep{Xing.2018}. Hohmann et al. showed in their work that high-strength materials ($R_{\mathrm{m}} >$ 600 MPa) tend to abrasive wear on the tip while soft graded steels  ($R_\mathrm{m} <$ 350 MPa) tend to adhesive wear on the lateral tool surface \citep{Hohmann.2017}. Many authors have shown that abrasive wear directly influences product quality in terms of cutting edge surface and burr height of the blanked workpieces \citep{Hambli2001FEM,Klingenberg2004}. They have carried out experimental and numerical studies on the correlation between cutting edge quality and abrasive wear state on the tool. One of the first studies in this area was conducted by Meade and Matsuno, who investigated the influence of wear on the burr height \citep{Maeda1967}. Cheung et al. explored the influence of different process variables on the wear state of the blanking tool and correlated it with  burr height and the required blanking force \citep{Cheung2000}. Kubik et al. systematically investigated the influence of semi-finished products and tool parameters on the quality of the cutting edge surface. They showed that the rounding of the cutting edge significantly increases the burr height and enhances the fraction of the sheared surface \citep{Kubik.2021}. 
In addition to experimental investigations, numerical methods are increasingly used to predict tool wear in blanking processes. Hambli implemented a finite element (FE) wear model to predict tool wear in blanking steel and correlated tool wear with burr formation \citep{Hambli2001FEM}. A more advanced approach was developed by Cheong and Kim, who took the effect of progressive tool wear into account of a FE model using a modified Archard wear model and a Lemaitre damage model. In addition, the geometry of the tool was adjusted based on the predicted wear volume, the tool replacement time was predicted, and the product quality was quantified based on the decrement of the hole area \citep{Cheon2016}.
Due to the non-linearity of the blanking process and the variety of process variables, a detailed description of the wear state by analytical, empirical or numerical models is only possible to a limited extent. In this context, data-driven approaches based on acquired process data offer the possibility to identify the cause of wear as well as to quantify the extent of the worn out region.

\begin{figure}[htb]
	\centering
	\fontsize{9pt}{9pt}\selectfont
	\def\svgwidth{\textwidth}
	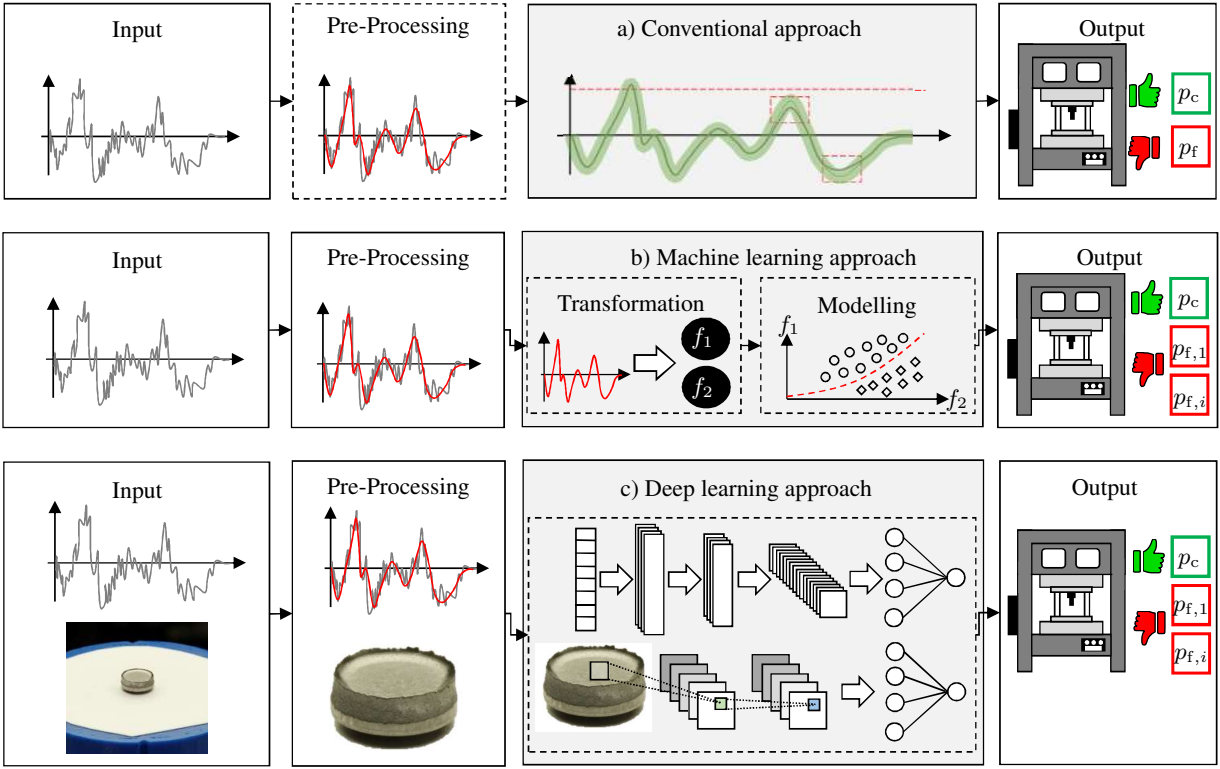
	\caption{Overview over data driven monitoring approaches}
	\label{Monitoringapproaches}
\end{figure}

\subsection{Data driven methods for process monitoring}
Since the 1990s, data-driven monitoring approaches have been researched and used in blanking processes. The approaches, which are mostly based on time series, differ considerably. Conventional monitoring approaches define envelope curves or thresholds that enable time series or extracted features to be monitored, whereby only binary statements about faultless $p_\mathrm{c}$ or faulty process states $p_\mathrm{f}$ can be made. More advanced approaches use machine learning algorithms that offer the potential to detect different error patterns $p_{\mathrm{f,}i}$. Recent publications in other production processes use deep learning algorithms, which can deal with a variety of raw data. Additionally there is no need for hand-crafted feature engineering, extraction and selection. An overview of the abstracted processes of the different approaches is provided in Fig. \ref{Monitoringapproaches}.

\subsubsection{Conventional monitoring approaches}
One of the first qualitative investigations on time series-based monitoring of blanking processes go back to Breitling et al. \citep{breitling1997process}, who integrated different sensors both in the press frame and in the progressive die. They showed that the optical appearance of the signals differs when using different materials, stroke speeds and punch-die clearances, from which the fundamental suitability of monitoring blanking processes through time series is justified. The first real-time monitoring of tool wear in a blanking process was published by Lee et al. \citep{Lee.1997} who used force signals to build an autoregressive model in order to estimate blanking force peaks. The coefficients of the autoregressive model were used as inputs for a linear discriminant function that classifies between new and worn tool. Koh et al. showed how wavelet transform-based Haar coefficients from a force signal can be used to detect faulty conditions in a stamping process \citep{Koh.1999}, similar wavelet-based approaches were investigated on force \citep{Jin.1999,Jin.2001,zhou2006spc} and acceleration signals \citep{Ge.2002}. Zhang et al. applied bispectral analysis to acceleration signals to identify defective products \citep{zhang2002bispectral}. Recent publications monitor stamping processes by recording acoustic emissions. Ubhayaratne et al. revealed correlations between wear and lubricant conditions with features of acoustic emission signals in the time \citep{ubhayaratne2015audio} and frequency domain \citep{ubhayaratne2017audio}. The potentials of applying time-frequency transformations to acoustic signals, such as the short-time Fourier transform, wavelet packet transform and Hilbert-Huang transform, were presented by Shanbhag et al. \citep{shanbhag2018investigating,shanbhag2020investigation}.

\subsubsection{Machine learning approaches}
The use of machine learning algorithms to monitor production processes opens up the possibility of classifying conditions in a much more detailed way or predicting quantitatively labels using regression algorithms. For this reason, machine learning algorithms are widely used in academic research as well as in industrial production, e.g., for condition monitoring or predictive maintenance.
\\The first application of a machine learning algorithm in a stamping process goes back to Jin and Shi, who used principal component analysis (PCA) to extract features from force signals. The PCA features were used as input variables for a decision tree to classify the combined presence of different process states, such as material thickness, lubrication, stroke speed and blank holder pressure \citep{Jin.2000}. Ge et al. extracted coefficients of an autoregressive model from the force signal of a stamping process and fed them into a hidden Markov model. Their approach can distinguish different fault types such as deviations from the desired material thickness, misfeed or slug with 87.5 \% classification accuracy \citep{Ge.2004b}. In another publication, the authors applied support vector machines and showed that accuracies of almost 100 \% can be achieved \citep{Ge.2004}.
\\Unsupervised learning algorithms are also used in stamping processes. Bassiuny et al. applied empirical mode decomposition to force signals of a stamping process and obtained intrinsic mode functions from which they extracted the signal energy and the Hilbert marginal spectrum as features. Then, they used learning vector quantisation to detect misfeeds and sheet thickness deviations \citep{Bassiuny.2007}. Ge et al. used a self-organizing map to classify five different fault types \citep{ge2003fault}. Bergs et al. recorded force signals in a blanking process, extracted 14 statistical features from the time domain and applied PCA to them.  The Euclidean and Mahalanobis distance of the first two PCA components were used to detect anomalies in consecutive punches. The authors showed that the process is subject to extensive fluctuations even with constant process settings \citep{Bergs.12020}.
\subsubsection{Deep learning approaches}
The application of machine learning algorithms to time signals is accompanied by the application of transformations, hand-crafted feature engineering, extraction and selection. Features must be extracted either model-based, e.g., by autoregressive models or PCA, in the time, frequency or time-frequency domain \citep{wang2006condition}. Deep learning approaches tackle this problem by unifying the process steps of data transformation and model building, consequently expanding the variety of usable data \citep{lecun-bengio-95a}. For example, the application of filtering and pooling operations in CNNs can simplify the high-dimensional representation of image data so that it can be used efficiently for classification and regression tasks. Publications from stamping, blanking and in general sheet metal forming processes rarely apply deep learning algorithms. Huang and Dzulfikri transformed acceleration signals into the frequency domain and used the entire power density spectrum as input for a one dimensional CNN. They demonstrated that they can distinguish seven different wear states with more than 99 \% accuracy \citep{Huang.2021}. Unterberg et al. recorded magnetic barkhausen noise from different material coils, created recurrence plots of the time series and used them as inputs for a two dimensional CNN. They classified whether certain material sections belong to the beginning, middle or end of a coil, as these sections have different semi-finished product properties \citep{unterberg2019situ}.
\\Approaches that use deep learning for the purpose of tool condition monitoring are relatively new. This is evidenced by the fact that a comprehensive review paper from 2013 did not present any deep learning applications \citep{dutta2013application}, whereas recent review papers discussed and emphasised the suitability of such algorithms \citep{serin2020review}. For example, Gouarir et al. recorded triaxial force signals in a milling process, applied a method called Gramian Angular Summation Field to the time series to obtain a two-dimensional representation and classified the force signals according to three wear states using CNN \citep{gouarir2018process}. Cao et al., Kothuru et al. and Martínez-Arellano et al. also classified tool wear conditions by transforming recorded time series into two-dimensional representations, which are then used as inputs for a CNN \citep{cao2019intelligent,kothuru2019application,martinez2019tool}. Recent publications used image data in combination with deep learning algorithms to detect wear states. A particularly interesting approach was taken by Bergs et al. who took 50 microscopic images of each of eight different tools. Then, they generated artificial images using data augmentation techniques and fed the images into a Fully Convolutional Network (FCN). The task of the FCN was to separate worn-out areas of the tool from non-worn-out areas, so that each pixel was labeled according to whether it represents a worn area or not. The authors showed that the deep learning approach is superior to classical computer vision algorithms and that local wear can be detected using FCN \citep{bergs2020digital}. Marei et al. took 327 microscope images of the cutting tool flank of a CNC machine and applied different CNN transfer learning models to them to classify and regress 14 different wear states of the tool. The ResNet-18 model led to both the highest classification accuracy of 84 \% and the lowest normalized mean absolute error of 0.0773 \citep{marei2021transfer}. Other image-based CNN approaches can be found in the literature, such as predicting the remaining useful life of cutting wheels \citep{li2020industrial} or classifying different types of wear in a face milling process \citep{wu2019automatic}.
These findings prove that image-based methods are highly suitable for the detection of wear states. Since wear conditions are reflected in the geometric properties of the products, especially in blanking processes \citep{hambli2002relationships,Kubik.2021,hambli2003numerical,mucha2010experimental}, the authors of the present paper choose the approach of taking images of workpieces and correlating them with the existing tool wear using CNN.
\subsection{Convolutional Neural Networks}
CNNs are one of the most important deep learning algorithms and are used in different application areas, such as object detection, object recognition, pose estimation and text recognition \citep{liu2017survey}. Ever since CNN showed unbeatable performance in high-level image classification competitions compared to shallow learning algorithms \citep{krizhevsky2012imagenet}, they have become a standard in image classification applications.
\begin{figure*}[h!]
	\centering
	\fontsize{9pt}{9pt}\selectfont
	\def\svgwidth{\textwidth}
\begingroup%
  \makeatletter%
  \providecommand\color[2][]{%
    \errmessage{(Inkscape) Color is used for the text in Inkscape, but the package 'color.sty' is not loaded}%
    \renewcommand\color[2][]{}%
  }%
  \providecommand\transparent[1]{%
    \errmessage{(Inkscape) Transparency is used (non-zero) for the text in Inkscape, but the package 'transparent.sty' is not loaded}%
    \renewcommand\transparent[1]{}%
  }%
  \providecommand\rotatebox[2]{#2}%
  \newcommand*\fsize{\dimexpr\f@size pt\relax}%
  \newcommand*\lineheight[1]{\fontsize{\fsize}{#1\fsize}\selectfont}%
  \ifx\svgwidth\undefined%
    \setlength{\unitlength}{654.83331299bp}%
    \ifx\svgscale\undefined%
      \relax%
    \else%
      \setlength{\unitlength}{\unitlength * \real{\svgscale}}%
    \fi%
  \else%
    \setlength{\unitlength}{\svgwidth}%
  \fi%
  \global\let\svgwidth\undefined%
  \global\let\svgscale\undefined%
  \makeatother%
  \begin{picture}(1,0.27278239)%
    \lineheight{1}%
    \setlength\tabcolsep{0pt}%
    \put(0,0){\includegraphics[width=\unitlength,page=1]{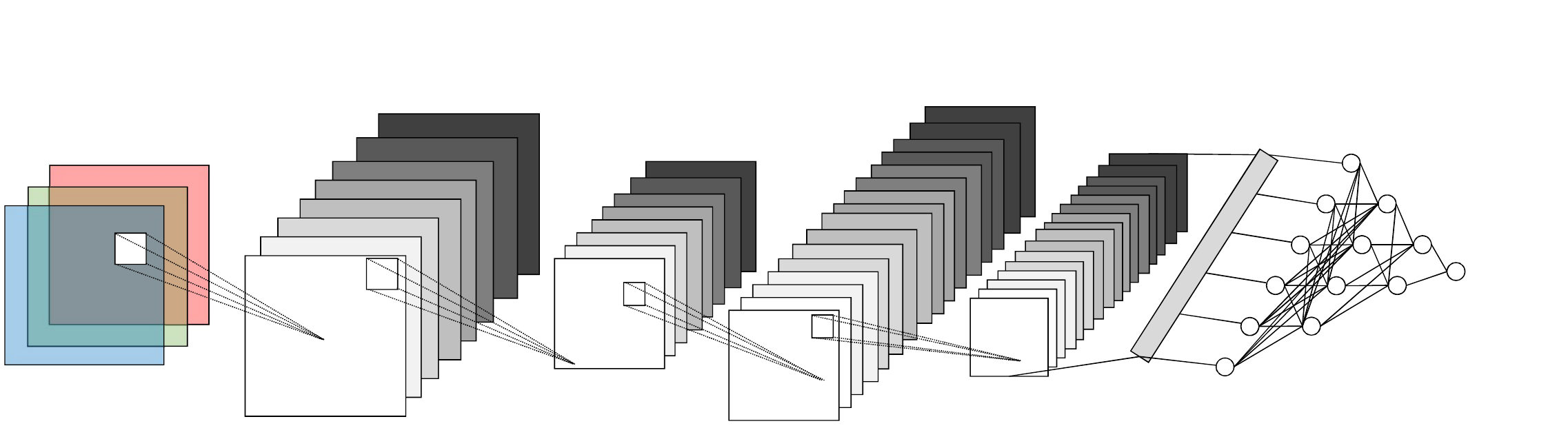}}%
    \put(0.08073422,0.2460782){\makebox(0,0)[lt]{\lineheight{1.25}\smash{\begin{tabular}[t]{l}Input \\Layer\end{tabular}}}}%
    \put(0.26648538,0.24567851){\makebox(0,0)[lt]{\lineheight{1.25}\smash{\begin{tabular}[t]{l}Convolutional \\Layer\end{tabular}}}}%
    \put(0.44910821,0.24567851){\makebox(0,0)[lt]{\lineheight{1.25}\smash{\begin{tabular}[t]{l}Pooling \\Layer\end{tabular}}}}%
    \put(0.80941532,0.24707183){\makebox(0,0)[t]{\lineheight{1.25}\smash{\begin{tabular}[t]{c}Flatten \\Layer\end{tabular}}}}%
    \put(0.84388321,0.04123476){\makebox(0,0)[lt]{\lineheight{1.25}\smash{\begin{tabular}[t]{l}Fully Connected \\Layer\end{tabular}}}}%
    \put(0.9330724,0.1693136){\makebox(0,0)[lt]{\lineheight{1.25}\smash{\begin{tabular}[t]{l}Output \\Layer\end{tabular}}}}%
    \put(0.03796919,0.1870681){\makebox(0,0)[lt]{\lineheight{1.25}\smash{\begin{tabular}[t]{l}$\textbf{X}_\mathrm{in}\in\mathbb{R}^{m \times m \times 3}$\end{tabular}}}}%
  \end{picture}%
\endgroup%

	\caption{Structure of a CNN}
	\label{StructureCNN}
\end{figure*}
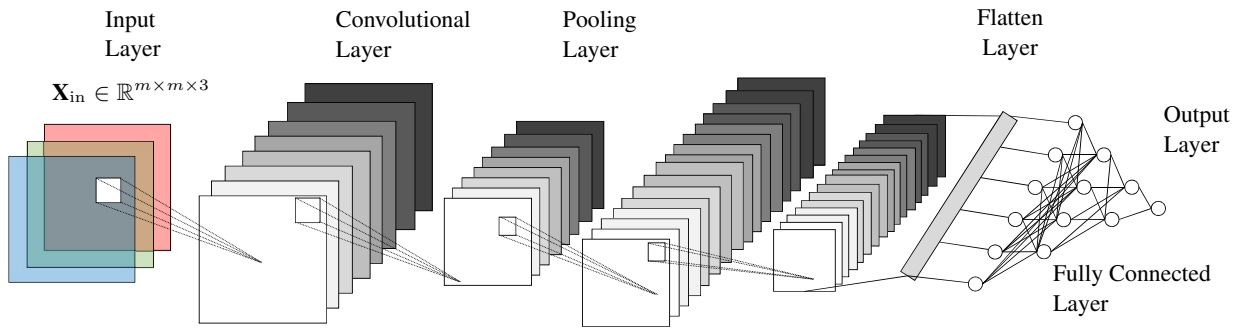
\\CNNs consist of an input layer and an output layer, between which convolutional, pooling, flatten and, if required, fully connected layers are arranged. The typical structure of a CNN is shown in Figure \ref{StructureCNN}. The input variable of image-processing CNNs are the original images, which can be represented as a two-dimensional matrix $\textit{\textbf{X}}_\mathrm{in}$ in the case of black-and-white images, or as a three-dimensional tensor $\textbf{X}_\mathrm{in}$ in the case of RGB images. One of the most important components of CNNs are convolutional layers. In convolutional layers, feature maps are generated by applying filters, also called kernels. If $\textit{\textbf{X}}_j^l$ is the $j$-th feature map in the $l$-th layer, it is calculated according to 
\begin{equation}
	\textit{\textbf{X}}_j^l=\textit{\textbf{f}}\left(\sum_{i=1}^N\textit{\textbf{X}}_i^{l-1}*\textit{\textbf{W}}_{ij}^l+\textit{\textbf{b}}_j^l\right),
\end{equation}
where $\textit{\textbf{W}}_{ij}^l$ is a weighting matrix, $*$ is the computational symbol of a two dimensional convolution with predefined filters and $\textit{\textbf{b}}_j^l$ are bias parameters of the l-th layer. Here, $\textit{\textbf{f}}(\cdot)$ is a non-linear activation function, whereby especially in image processing deep networks Rectified Linear Unit (ReLU) 
\begin{equation}
	f(x)=\mathrm{max}[0,x]
\end{equation}
is used. The convolutional layer is usually followed by pooling layers, which reduce the dimensionality of the previous feature map according to predefined rules. In common CNNs, average pooling or max pooling layers are used. After a mostly alternating sequence of convolutional and pooling layers, a flatten layer follows, which transforms the multi-dimensional feature map into a one dimensional representation. At the end of the network topology, several fully connected layers usually form a multilayer perceptron (MLP), which transforms the one dimensional input to output values. Within the MLP, the output of the $l$-th hidden layers is calculated by
\begin{equation}
	\textit{\textbf{x}}^l=\textit{\textbf{f}}\left( \textit{\textbf{W}}^l\cdot \textit{\textbf{x}}^{l-1}+\textit{\textbf{b}}^l\right).
\end{equation}
In the classification case, the output layer has as many neurons as there are classes, with the softmax function
\begin{equation}
	f_{\mathrm{soft,}k}(x_i)=\frac{e^{x_k}}{\sum_{i=1}^Ne^{x_i}}
\end{equation}
now serving as the activation function of the $k$-th output neuron.\\ 
\begin{wrapfigure}[24]{o}{0.48\textwidth}
	\centering
	\fontsize{9pt}{9pt}\selectfont
	\def\svgwidth{0.48\textwidth}
	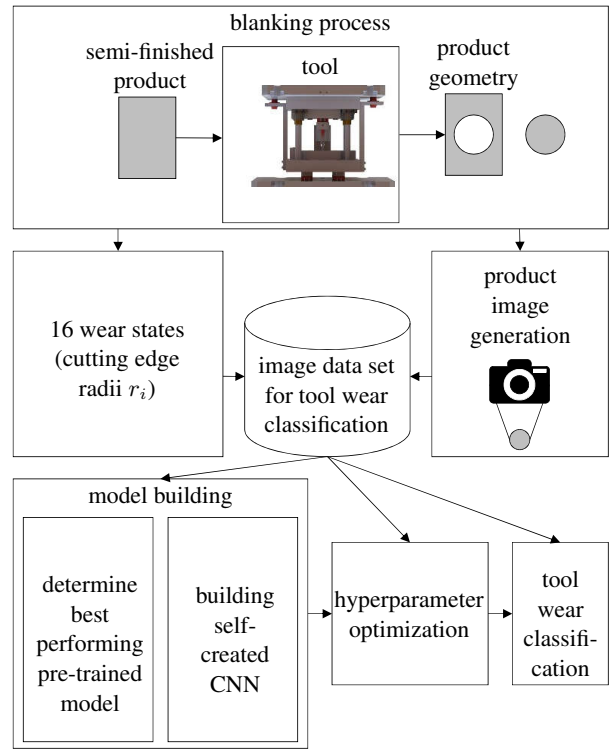
	\caption{Procedure for classifying tool wear conditions in a blanking process using image-based CNN}
	\label{OverviewProcedure}
\end{wrapfigure}
\section{Methodology}
\label{sec:others}
\begin{figure}{h!}
	\centering
	\fontsize{9pt}{9pt}\selectfont
	\def\svgwidth{0.4\columnwidth}
\begingroup%
  \makeatletter%
  \providecommand\color[2][]{%
    \errmessage{(Inkscape) Color is used for the text in Inkscape, but the package 'color.sty' is not loaded}%
    \renewcommand\color[2][]{}%
  }%
  \providecommand\transparent[1]{%
    \errmessage{(Inkscape) Transparency is used (non-zero) for the text in Inkscape, but the package 'transparent.sty' is not loaded}%
    \renewcommand\transparent[1]{}%
  }%
  \providecommand\rotatebox[2]{#2}%
  \newcommand*\fsize{\dimexpr\f@size pt\relax}%
  \newcommand*\lineheight[1]{\fontsize{\fsize}{#1\fsize}\selectfont}%
  \ifx\svgwidth\undefined%
    \setlength{\unitlength}{211.40210724bp}%
    \ifx\svgscale\undefined%
      \relax%
    \else%
      \setlength{\unitlength}{\unitlength * \real{\svgscale}}%
    \fi%
  \else%
    \setlength{\unitlength}{\svgwidth}%
  \fi%
  \global\let\svgwidth\undefined%
  \global\let\svgscale\undefined%
  \makeatother%
  \begin{picture}(1,1.21985538)%
    \lineheight{1}%
    \setlength\tabcolsep{0pt}%
    \put(0,0){\includegraphics[width=\unitlength,page=1]{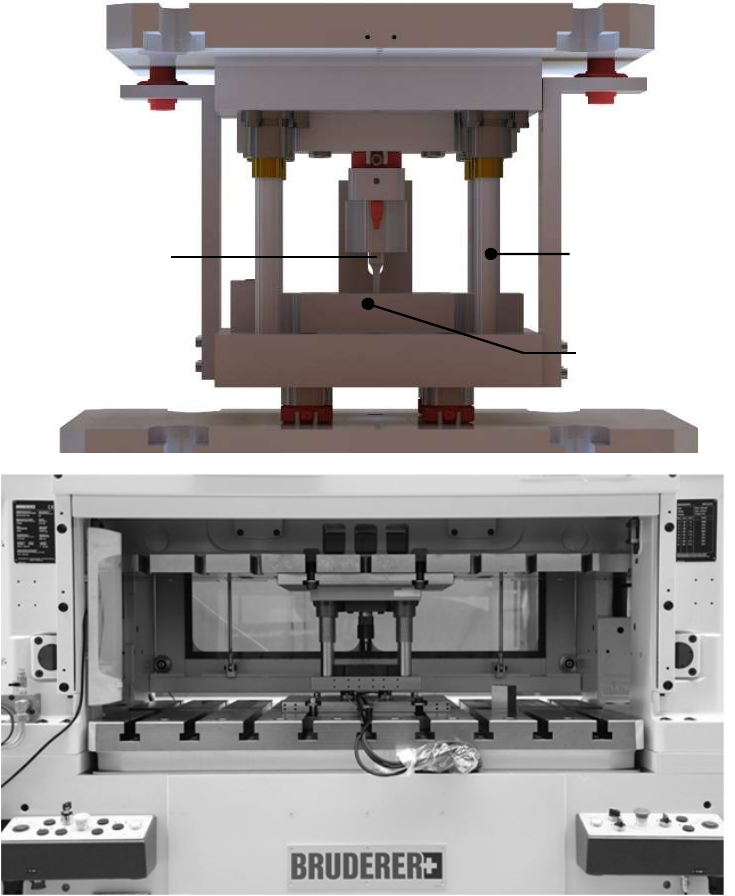}}%
    \put(0.1241656,0.85624375){\makebox(0,0)[lt]{\lineheight{1.14999998}\smash{\begin{tabular}[t]{l}punch\\\\\end{tabular}}}}%
    \put(0.78916857,0.91549971){\makebox(0,0)[lt]{\lineheight{1.25}\smash{\begin{tabular}[t]{l}\\guidance\\\end{tabular}}}}%
    \put(0,0){\includegraphics[width=\unitlength,page=2]{Fig4.pdf}}%
    \put(0.8245074,0.72117003){\makebox(0,0)[t]{\lineheight{1.25}\smash{\begin{tabular}[t]{c}die\end{tabular}}}}%
    \put(0.00660911,1.17616126){\makebox(0,0)[lt]{\lineheight{1.14999998}\smash{\begin{tabular}[t]{l}a)\\\\\end{tabular}}}}%
    \put(0.00514449,0.58828262){\makebox(0,0)[lt]{\lineheight{1.14999998}\smash{\begin{tabular}[t]{l}b)\\\\\end{tabular}}}}%
  \end{picture}%
\endgroup%

	\caption{Setup of the blanking tool (a) and blanking machine (b)}
	\label{fig:experimental_setup}
\end{figure}
In order to quantify the state of abrasive wear based on images of the blanked workpieces, a CNN is used to classify sixteen wear states of the cutting edge radii $r_i$. The applied procedure is shown in Figure \ref{OverviewProcedure}. In the first step, experiments are carried out and 1860 workpieces are manufactured. Subsequently, the photos are taken, which form a labeled image data set for tool wear classification by assigning the individual cutting edge radii $r_i$. In the model building step, on the one hand, it is investigated which pre-trained image classification model is suitable for wear detection, on the other hand, a self-created model topology is used. The models resulting from the different approaches are hyperparameter optimized and finally their performance is quantified and compared with each other.

\subsection{Experimental setup}
The experiments to produce the blanked workpieces are performed on a mechanical high-speed press from Bruderer AG (BSTA 810). The machine has a nominal force of 810 \,kN and stroke rates of up to 1000 spm at a stroke height of 16 mm. The experiments are carried out with a stroke speed of 300 spm and a stroke distance of 35 mm. As an experimental material a micro-alloyed steels with high yield strengths for cold forming approaches (1.0480) is used. Figure \ref{fig:experimental_setup} shows the experimental set-up, including the blanking tool. Table \ref{ParameterTable} summarizes the selected parameters of the press and the tool as well as the properties of the experimental material.

\begin{wraptable}{o}{5cm}
	\centering
	\fontsize{9pt}{9pt}\selectfont
	\caption{Selected parameters of the process and material}
	\label{ParameterTable}
	\begin{tabular}{c c}
		\hline
		\multicolumn{2}{c}{process parameters}\\
		\hline
		stroke speed & 300 spm\\
		stroke distance & $35$ mm\\
		clearance & 7.5 \% \\
		\hline
		\multicolumn{2}{c}{material parameters}\\ 
		\hline
		description & 1.0480 \\
		tensile strength &  $365  \mathrm{N/mm^2}$\\
		elongation $\mathrm{A_{80}}$ & 27.10 \%\\
		sheet thickness & $(2.00\pm0.1)$ mm\\
		\hline
	\end{tabular}
\end{wraptable}

\subsection{Data generation}\label{Data Generation}

\begin{wrapfigure}[19]{0}{0.5\textwidth}
	\centering
	\fontsize{8pt}{9pt}\selectfont
	\def\svgwidth{0.45\columnwidth}
\begingroup%
  \makeatletter%
  \providecommand\color[2][]{%
    \errmessage{(Inkscape) Color is used for the text in Inkscape, but the package 'color.sty' is not loaded}%
    \renewcommand\color[2][]{}%
  }%
  \providecommand\transparent[1]{%
    \errmessage{(Inkscape) Transparency is used (non-zero) for the text in Inkscape, but the package 'transparent.sty' is not loaded}%
    \renewcommand\transparent[1]{}%
  }%
  \providecommand\rotatebox[2]{#2}%
  \newcommand*\fsize{\dimexpr\f@size pt\relax}%
  \newcommand*\lineheight[1]{\fontsize{\fsize}{#1\fsize}\selectfont}%
  \ifx\svgwidth\undefined%
    \setlength{\unitlength}{246.71999359bp}%
    \ifx\svgscale\undefined%
      \relax%
    \else%
      \setlength{\unitlength}{\unitlength * \real{\svgscale}}%
    \fi%
  \else%
    \setlength{\unitlength}{\svgwidth}%
  \fi%
  \global\let\svgwidth\undefined%
  \global\let\svgscale\undefined%
  \makeatother%
  \begin{picture}(1,0.87062248)%
    \lineheight{1}%
    \setlength\tabcolsep{0pt}%
    \put(0,0){\includegraphics[width=\unitlength,page=1]{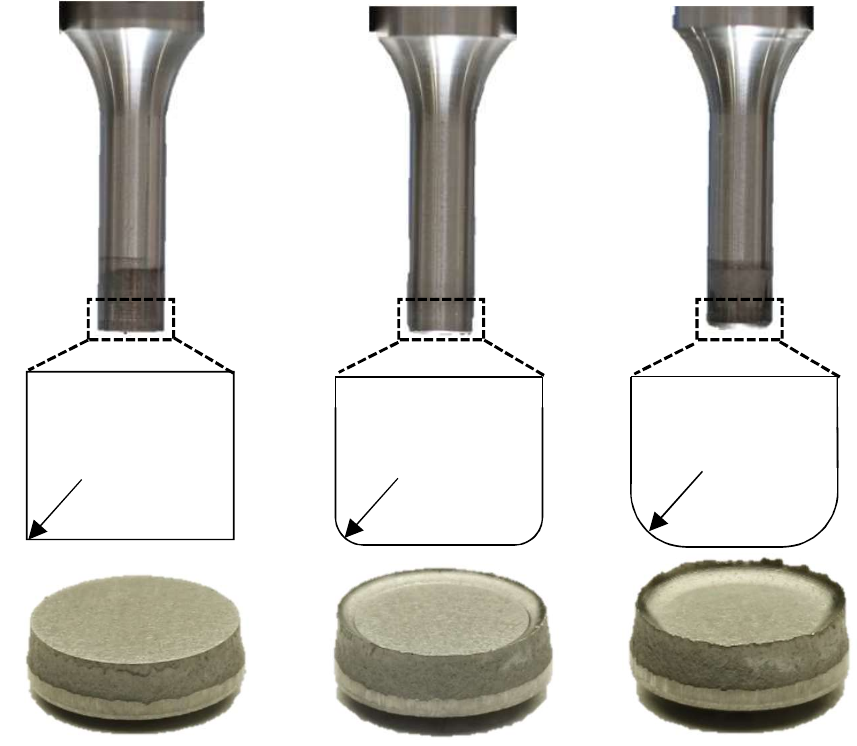}}%
    \put(0.14892664,0.34243502){\color[rgb]{0,0,0}\makebox(0,0)[t]{\lineheight{1.25}\smash{\begin{tabular}[t]{c}$r_\mathrm{00}= 0$ mm\end{tabular}}}}%
    \put(0.51166344,0.34068649){\color[rgb]{0,0,0}\makebox(0,0)[t]{\lineheight{1.25}\smash{\begin{tabular}[t]{c}$r_\mathrm{40}= 0.4$ mm\end{tabular}}}}%
    \put(0.85794111,0.34243502){\color[rgb]{0,0,0}\makebox(0,0)[t]{\lineheight{1.25}\smash{\begin{tabular}[t]{c}$r_\mathrm{80}= 0.8$ mm\end{tabular}}}}%
    \put(0.01075119,0.83394848){\color[rgb]{0,0,0}\makebox(0,0)[lt]{\lineheight{1.25}\smash{\begin{tabular}[t]{l}a)\end{tabular}}}}%
    \put(0.37453528,0.83394848){\color[rgb]{0,0,0}\makebox(0,0)[lt]{\lineheight{1.25}\smash{\begin{tabular}[t]{l}b)\end{tabular}}}}%
    \put(0.72948637,0.83394848){\color[rgb]{0,0,0}\makebox(0,0)[lt]{\lineheight{1.25}\smash{\begin{tabular}[t]{l}c)\end{tabular}}}}%
  \end{picture}%
\endgroup%

	\caption{Workpieces produced with cutting edge radii $r_i$ of a) 0 mm, b) 0.4 mm and c) 0.8 mm}
	\label{Bauteile}
\end{wrapfigure}
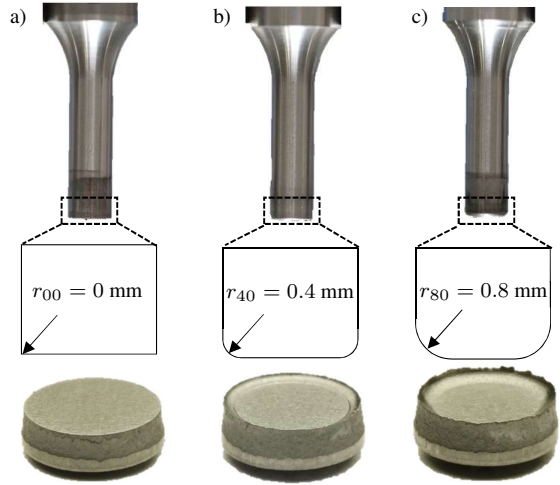

As input of the CNN, a comprehensive data set is generated with 7440 workpiece images. When generating the data set, two basic principles must be taken into consideration. On the one hand, product properties that correlate with tool wear must be represented as best as possible, and on the other hand, the images should be reproducible and taken from the same perspective. This prevents the model from suffering performance degradation as it has to generalize more than necessary.
\\To meet these criteria, the images are shot in a illuminated environment using a rigid camera mount from an oblique side view. Each of the 1860 workpieces is photographed four times, with the images taken 90 degrees apart from each other. The camera used is a Canon EOS 6D SLR, which is capable of capturing images at $3648 \times 3648$ pixels. The images are then cropped to the region of interest consisting of $800 \times 800$ pixels and finally resized to the target input size of the corresponding CNN.  Figure \ref{Bauteile} shows three workpieces that have been produced with punches with different levels of cutting edge radii $r_i$.
As the used Deep Learning approach is based on a supervised learning technique, the generated image data set is labeled with the wear states of the punch. It is assumed that abrasive wear causes the rounding of the cutting edge radii of the punch \citep{Kubik.2021,hambli2002prediction}. In order to replicate the desired abrasive wear conditions without the need for time-consuming long-term experiments, the cutting edges are mechanically rounded by a post machining process. The cutting edge radii $r_i$ are varied in sixteen steps according to
\begin{equation}
	r_i \in \{0, 0.10, 0.15, 0.20, 0.25, 0.30, 0.35, 0.40, 0.45,\\ 0.50, 0.55, 0.60, 0.65, 0.70, 0.75, 0.80\} \mathrm{mm}.
\end{equation}
After the images are taken, they are read into Python and then divided into training, validation and test data. Care was taken to ensure that images of the same workpiece are only used in the training or test data set, so that the test data set only consists of images of workpieces that are not known to the model. The RGB images are transformed into NumPy arrays so that the images are available as a tensor $\textbf{X}_\mathrm{in}$. Finally, the tensor elements are normalized to values between zero and one, since CNNs can best process this value range. 

\subsection{Deep Learning Modeling}
In order to find the best possible network topology, two different approaches are chosen. On the one hand, network topologies can be freely chosen and then hyperparameter optimized. On the other hand, transfer learning techniques \citep{zhuang2020comprehensive} allow the use of pre-trained models whose performance has already been proven in other image classification tasks \citep{shao2014transfer}. The machine and deep learning community was able to demonstrate several years ago that CNNs are capable of learning generic mid-level image representations \citep{oquab2014learning}. As a result, weights and topologies of networks obtained in image classification tasks with millions of labeled images and several thousand classes can also bring advantages to  tasks that have comparatively few images and classes \citep{sermanet2013overfeat}.

\subsubsection{Choice of the transfer learning model}
\begin{figure}[t!]
	\centering
	\fontsize{9pt}{9pt}\selectfont
	\def\svgwidth{\textwidth}
	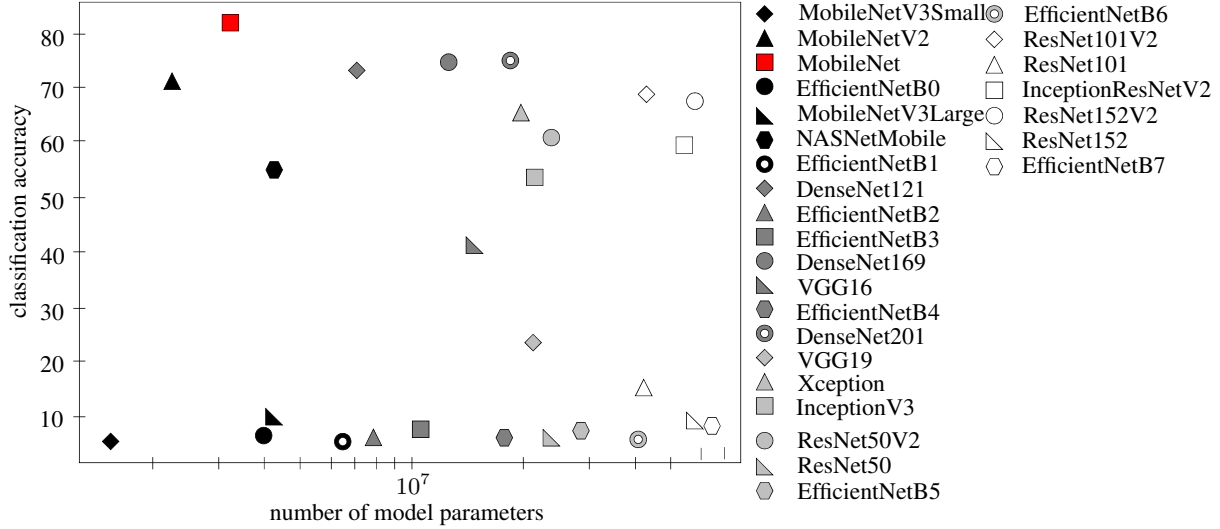
	\caption{Comparison of pretrained models}
	\label{Comparisonpretrained}
\end{figure}

All models presented are developed in Python, popular deep learning libraries such as Keras and Tensorflow \citep{abadi2016tensorflow} are used. Keras provides a variety of pre-trained models that have been used for the transfer learning approach. The dataset described in chapter \ref{Data Generation} is divided into 70 \% test, 15 \% validation and 15 \% training data and fed into 27 pre-trained models. Sparse categorial cross entropy (SCCE) is used as the loss function and Adam \citep{kingma2014adam} as the optimization algorithm. No additional fully connected layers are added to the models, the inputs of the output layer are taken from the flatten layer and the training of the models is stopped after 5 epochs.  The results of the comparison of all pretrained models can be seen in Figure \ref{Comparisonpretrained}.
\begin{table}[h!]
	\centering
	\caption{Overview over optimized hyperparameters}
	\label{Hyperparametertable}
	\begin{tabular}{c c c}
		\hline
		Hyperparameter & MobileNet & Self-created CNN \\
		\hline
		Amount of FCLs & Not used (0) & 3 \\
		Neurons in FCLs & Not used & 512;\,512;\,512\\
		Batch Normalization & Not used & After every FCL\\
		Dropout & Not used & Not used\\
		L2-regularization & Not used & 0.01\\
		Amount of CL & Not optimized & 7\\
		Learning Rate & 0.0001 & 0.001\\
		Trained layers & Last 24 layers & All layers\\
		Filters in CL & Not optimized & $2^5$;$2^6$;$2^6$;$2^7$;$2^8$;$2^8$;$2^8$\\
		\hline
	\end{tabular}
\end{table}

\begin{wraptable}{o}{9cm}
	\caption{Overview over network topologies and informations on the training process}
	\label{NetworkTopology}
	\begin{tabular}{c c c}
		\hline
		Network parameter & MobileNet & Self-created CNN \\
		\hline
		Activation function & ReLU & ReLU \\
		Amount of layers & 88 & 31\\
		Trainable parameters & 3,223,376 & 2,273,680\\
		Output activation function & Softmax & Softmax\\
		Pooling operation used & Average & Max\\
		Amount of CL & 27 & 7\\
		Amount of PL & 1 & 6\\
		Input dimension of $\textbf{X}_\mathrm{in}$ & $\mathbb{R}^{224 \times224 \times 3}$ & $\mathbb{R}^{128 \times 128 \times 3}$\\
		Amount of output neurons & 16 & 16 \\
		Amount of FCL & 0 & 3\\
		Optimization algorithm & Adam & Adam\\
		Loss function & SCCE & SCCE\\
		\hline
	\end{tabular}
\end{wraptable}

It is evident that the accuracy of the different models varies greatly. The VGG models \citep{tammina2019transfer} deliver classification accuracies between 45 and 50 \%, whereas different versions of ResNet models achieve accuracies of up to 79 \%. The MobileNet, which is originally used for object recognition in commercially available smartphones \citep{howard2017mobilenets}, performs best with classification accuracies over 80 \%. Last but not least due to the comparatively low number of trainable parameters and the associated low model complexity, it was decided to use the MobileNet as transfer learning model in the following.

\subsubsection{Hyperparameter optimization}

In order to determine optimal hyperparameters for the developed models, a comprehensive hyperparameter optimization is carried out for both the self-created CNN and the pre-trained MobileNet. Due to the probability-based search and computational efficiency advantages over conventional optimization methods such as grid search, Bayesian hyperparameter optimization \citep{wu2019hyperparameter} is applied. In the optimization process, the individual runs are trained on 50 epochs and those hyperparameters are selected that lead to the highest accuracies. Table \ref{Hyperparametertable} gives an overview of the optimized hyperparameters. Table \ref{NetworkTopology} presents the final network topologies obtained and additional information on the training processes.

\section{Results}
\begin{wraptable}{o}{10 cm}
	\centering
	\caption{Descriptive statistics of the model performances}
	\label{DescriptiveStatistics}
	\begin{tabular}{c c c}
		\hline
		Statistical parameter & MobileNet & Self-created CNN \\
		\hline
		Mean accuracy $\mu$ & 98.94 \% & 98.80 \% \\
		Median accuracy & 98.93 \% & 98.83 \% \\
		Standard deviation $\sigma$ & 0.24 \% & 0.37 \% \\
		95\% confidence intervals & $(98.94 \pm 0.06) \%$ & $(98.80 \pm 0.10) \%$ \\
		Maximum accuracy & 99.41 \% & 99.71 \% \\
		Minimum accuracy & 98.44 \% & 97.47 \% \\
		\hline
	\end{tabular}
\end{wraptable}
In the present section, the performances of the two models obtained from hyperparameter optimization are compared. To be able to compare the model performances in a statistically robust way, the two models are both trained 100 times over 200 epochs, whereby the training process is stopped if no significant performance improvements are recorded over 30 consecutive epochs. 64 images of each wear class are used as validation data, another 64 images of each class as test data. The remaining 5392 images are used to train the models.
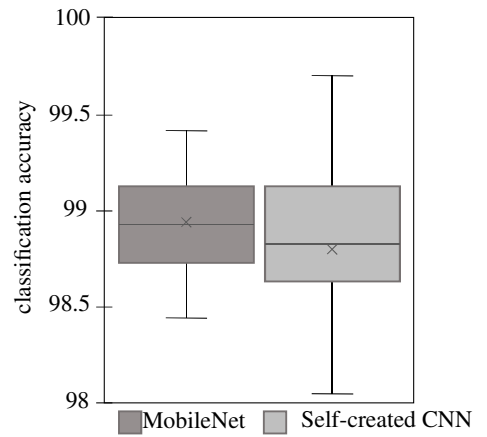
\begin{wrapfigure}[15]{r}{6cm}
	\fontsize{9pt}{9pt}\selectfont
	\def\svgwidth{0.35\columnwidth}
\begingroup%
  \makeatletter%
  \providecommand\color[2][]{%
    \errmessage{(Inkscape) Color is used for the text in Inkscape, but the package 'color.sty' is not loaded}%
    \renewcommand\color[2][]{}%
  }%
  \providecommand\transparent[1]{%
    \errmessage{(Inkscape) Transparency is used (non-zero) for the text in Inkscape, but the package 'transparent.sty' is not loaded}%
    \renewcommand\transparent[1]{}%
  }%
  \providecommand\rotatebox[2]{#2}%
  \newcommand*\fsize{\dimexpr\f@size pt\relax}%
  \newcommand*\lineheight[1]{\fontsize{\fsize}{#1\fsize}\selectfont}%
  \ifx\svgwidth\undefined%
    \setlength{\unitlength}{193.36953735bp}%
    \ifx\svgscale\undefined%
      \relax%
    \else%
      \setlength{\unitlength}{\unitlength * \real{\svgscale}}%
    \fi%
  \else%
    \setlength{\unitlength}{\svgwidth}%
  \fi%
  \global\let\svgwidth\undefined%
  \global\let\svgscale\undefined%
  \makeatother%
  \begin{picture}(1,0.96970111)%
    \lineheight{1}%
    \setlength\tabcolsep{0pt}%
    \put(0,0){\includegraphics[width=\unitlength,page=1]{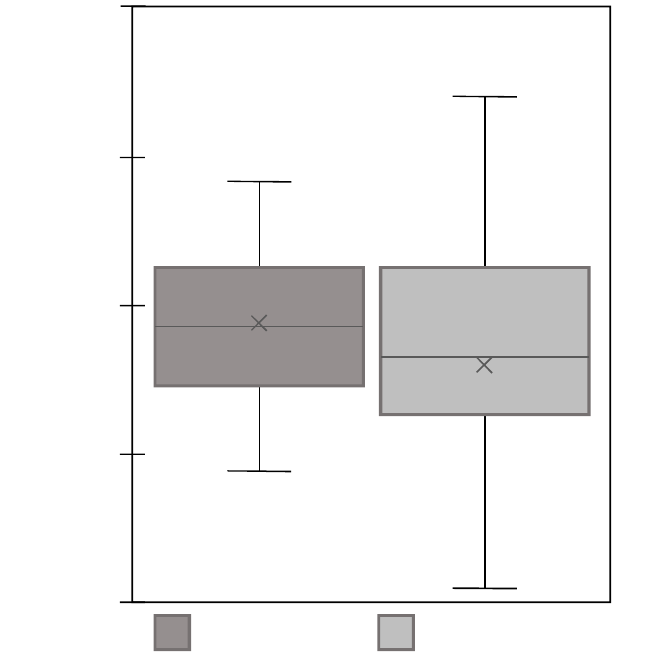}}%
    \put(0.1238279,0.94285151){\makebox(0,0)[t]{\lineheight{1.25}\smash{\begin{tabular}[t]{c}100\end{tabular}}}}%
    \put(0.11841057,0.72111872){\makebox(0,0)[t]{\lineheight{1.25}\smash{\begin{tabular}[t]{c}99.5\end{tabular}}}}%
    \put(0.13050641,0.50190487){\makebox(0,0)[t]{\lineheight{1.25}\smash{\begin{tabular}[t]{c}99\end{tabular}}}}%
    \put(0.11726196,0.28528012){\makebox(0,0)[t]{\lineheight{1.25}\smash{\begin{tabular}[t]{c}98.5\end{tabular}}}}%
    \put(0.1358062,0.06044428){\makebox(0,0)[t]{\lineheight{1.25}\smash{\begin{tabular}[t]{c}98\end{tabular}}}}%
    \put(0.40350461,0.01576055){\makebox(0,0)[t]{\lineheight{1.25}\smash{\begin{tabular}[t]{c}MobileNet\end{tabular}}}}%
    \put(0.84699799,0.0160786){\makebox(0,0)[t]{\lineheight{1.25}\smash{\begin{tabular}[t]{c}Self-created CNN\end{tabular}}}}%
    \put(0.02768037,0.5240377){\rotatebox{90}{\makebox(0,0)[t]{\lineheight{1.25}\smash{\begin{tabular}[t]{c}classification accuracy\end{tabular}}}}}%
  \end{picture}%
\endgroup%

	\caption{Performance comparison of MobileNet and self-created CNN}
	\label{Boxplot}
\end{wrapfigure}

Figure \ref{Boxplot} shows the results of 100 independent model runs as a boxplot, descriptive statistics of the results can be found in Table \ref{DescriptiveStatistics}. It can be seen that both models deliver extremely high accuracy values and the wear condition of the tool can be determined by both models with almost 99 \% accuracy. The accuracies of the models differ only in the decimal range, whereby the performance of the MobileNet is slightly better.  The robustness of the MobileNet also appears to be higher, which is indicated by lower standard deviations and closer confidence intervals of the model performance. This is either due to the initialised weights of the MobileNet and the associated ability to provide generic mid-level image representations, or to the higher number of trainable model parameters.

\begin{wraptable}{r}{7cm}
	\caption{Results of the one-tailed t-test for differences in mean accuracies between MobileNet and self-created CNN}
	\label{t-Test}
	\centering
	\begin{tabular}{c c}
		\hline
		Observations $N$ per model & 100 \\
		Degrees of freedom & 169 \\
		$t_\mathrm{stat}$  & 3.13 \\
		$t_\mathrm{crit}$ one-tailed & 2.35 \\
		$p(t_\mathrm{stat}\leq t_\mathrm{crit})$  & $0.001$ \\
		
		\hline
	\end{tabular}
\end{wraptable}

In order to make statistically reliable statements about the superiority of one machine learning algorithm over another, methods from statistics can be applied. One approach to detect significant differences in performances of classifiers is the Welch's unequal variances t-test for differences in mean accuracy \citep{demvsar2006statistical}. Since it is assumed that the pre-trained MobileNet can generate generic mid-level image representations, a one-tailed hypothesis test is conducted and the significance level is chosen to be $p=0.01$. If the mean accuracies of MobileNet and self-created CNN are described by $\mu_\mathrm{{Mob}}$ and $\mu_\mathrm{CNN}$, respectively, then the null hypothesis is:

$H_0$: The accuracy of the MobileNet is equal to or less than that of the self-created CNN 
$(\mu_\mathrm{Mob}\leq\mu_\mathrm{CNN})$.

The t-statistic value can be calculated by
\begin{equation}
	t_\mathrm{stat}=\frac{\mu_\mathrm{Mob}-\mu_\mathrm{CNN}}{\sqrt{\frac{\sigma_\mathrm{Mob}^2}{N}+\frac{\sigma_\mathrm{CNN}^2}{N}}},
\end{equation}
where $\sigma_\mathrm{Mob}$ and $\sigma_\mathrm{CNN}$ are the standard deviations of the accuracies of Mobilenet and self-generated CNN, respectively. The results of the t-test can be seen in Table \ref{t-Test}. It is shown that the test is  significant due to
\begin{equation}
	p(t_\mathrm{stat}\leq t_\mathrm{crit}) < p=0.01,
\end{equation}
which is why the null hypothesis must be rejected. This leads to the statistically robust statement that the MobileNet is better suited for the classification of tool wear conditions and leads to higher accuracies.

\begin{wrapfigure}{r}{0.5\textwidth}
	\centering
	\fontsize{9pt}{9pt}\selectfont
	\def\svgwidth{0.4\columnwidth}
	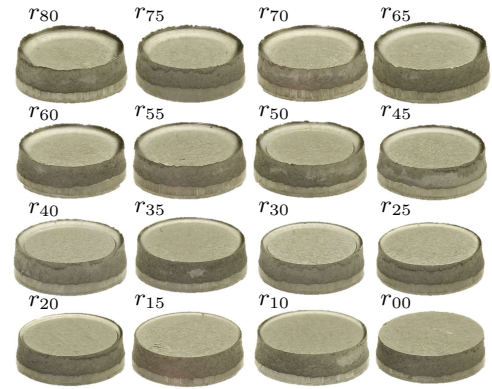
	\caption{Example images of workpieces produced with 16 different punch edge radii $r_i$}
	\label{OverviewProducts}
\end{wrapfigure}

\begin{table}[h!]
	\centering
	\caption{Cumulative confusion matrix for tool wear classification of all 100 trained MobileNets in \%}
	\label{Confusionmatrix}
	\fontsize{8}{10}\selectfont
	\begin{tabular}{ccccccccccccccccc}
		\hline
		$r_i$ & $\hat{r}_{00}$ & $\hat{r}_{10}$ & $\hat{r}_{15}$ & $\hat{r}_{20}$ & $\hat{r}_{25}$ & $\hat{r}_{30}$ & $\hat{r}_{35}$ & $\hat{r}_{40}$ & $\hat{r}_{45}$ & $\hat{r}_{50}$ & $\hat{r}_{55}$ & $\hat{r}_{60}$ & $\hat{r}_{65}$ & $\hat{r}_{70}$ & $\hat{r}_{75}$ & $\hat{r}_{80}$ \\
		\hline
		$r_{00}$ & 100 & - & - & - & - & - & - & - & - & - & - & - & - & - & - & - \\
		$r_{10}$ & - & 98.47 & .39 & - & 1.11 & - & .03 & - & - & - & - & - & - & - & - & - \\
		$r_{15}$ & - & - & 100 & - & - & - & - & - & - & - & - & - & - & - & - & - \\
		$r_{20}$ & - & - & .02 & 99.98 & - & - & - & - & - & - & - & - & - & - & - & -  \\
		$r_{25}$ & - & - & .05 & - & 99.95 & - & - & - & - & - & - & - & - & - & - & - \\
		$r_{30}$ & - & - & - & - & - & 100 & - & - & - & - & - & - & - & - & - & - \\
		$r_{35}$ & - & - & - & - & - & - & 100 & - & - & - & - & - & - & - & - & -  \\
		$r_{40}$ & - & - & - & - & - & - & - & 99.97 & - & .03 & - & - & - & - & - & - \\
		$r_{45}$ & - & - & - & - & - & - & - & - & 99.73 & - & .27 & - & - & - & - & - \\
		$r_{50}$ & - & - & - & - & - & - & - & - & - & 100 & - & - & - & - & - & -  \\
		$r_{55}$ & - & - & - & - & - & - & - & - & .02 & - & 98 & 1.98 & - & - & - & - \\
		$r_{60}$ & - & - & - & - & - & - & - & - & - & - & .52 & 96.86 & .22 & 2.41 & - & - \\
		$r_{65}$ & - & - & - & - & - & - & - & - & - & - & .86 & .81 & 98.23 & .08 & .02 & - \\
		$r_{70}$ & - & - & - & - & - & - & - & - & - & - & - & 1.03 & 2.05 & 94.67 & 1.89 & .36 \\
		$r_{75}$ & - & - & - & - & - & - & - & - & - & - & - & - & - & .03 & 99.97 & -  \\
		$r_{80}$ & - & - & - & - & - & - & - & - & - & - & - & - & - & - & 2.78 & 97.22 \\
		\hline
	\end{tabular}
\end{table}
Table \ref{Confusionmatrix} gives an overview over the percentage misclassifications of the 100 trained MobileNets in the form of a confusion matrix, from which the wear states that are difficult to classify for the models can be derived. While there are low numbers of misclassifications at low cutting edge radii and especially with unworn punches, the models seem to become less accurate with increasing wear. Figure \ref{OverviewProducts} shows example workpiece images for each wear class, which can be used to make assumptions about the increasing number of misclassifications with increasing wear. While differences in the geometric properties of the workpieces in the lower wear classes are particularly noticeable in the burr height and the impressions of the punch on the punching surface, which can thus potentially be discriminatory parameters for class classification, a differentiation of the high wear classes on the basis of optical properties is only possible to a limited extent. This is consistent with results from experimental studies in which the burr height grows asymptotically with increasing wear, reaches a plateau and therefore no longer serves as a discriminating feature above a certain degree of wear \citep{straffelini2010improving}. From wear class $r_\mathrm{55}$ upgoing, the punch leaves hardly any marks on the blanking surface. In addition, it can be seen that the burr height over the rotationally symmetrical edge shows strong variances, which make a local feature extraction for the CNN more difficult.
\\ Furthermore, the cutting surface also contains potentially discriminating features that cause a more accurate classification in lower wear classes. While especially in wear classes $\leq r_\mathrm{20}$ the share of the shear zone in the total height of the workpiece correlates positively with the wear, such tendencies are not recognizable in higher wear classes.

\section{Conclusion and outlook}

While monitoring approaches of blanking processes are nowadays largely based on time series, this paper presents an approach for image-based tool wear monitoring. For this purpose, abrasive wear is simulated by mechanically rounding the edges of punches, whereby 1860 workpieces are manufactured with 16 differently worn punches. By capturing 7440 workpiece images, a labeled data set is generated, which is then processed using Convolutional Neural Networks.
A pre-trained CNN called MobileNet as well as a self-generated CNN are hyperparameter optimized and used for the image classification task. The results show that both models have excellent classification accuracies of up to 99 \% and are thus highly suitable for tool wear classification. Using statistical testing methods, it can be shown that the pre-trained MobileNet is slightly superior to the self-created CNN. In addition, images of the workpieces show that optical properties correlate with the punch wear, but their ambiguity prevents an even more precise classification, especially in the presence of high wear.
\\These promising results open up new possibilities for future research. Future research efforts should investigate the extent to which models react to workpiece images of different semi-finished products and punch geometries. A highly interesting approach is to train the models on unlabeled data by using domain adaptation techniques in order to generate models that are as generalisable as possible and adaptable to different applications. Another interesting object of investigation is the performance behaviour of the models with varying data set sizes and how data augmentation techniques can improve model performances. In addition, the workpiece images offer the possibility to apply computer vision algorithms for hand-crafted feature extraction in order to capture product properties as time-efficiently as possible. A process-integrated acquisition of the workpiece images and their fusion with sensor data enables the identification of correlations between time series signals and product properties, which in turn can be used for predictive quality applications.

%
%

\bibliographystyle{spbasic}
\bibliography{Paper_CNN}  






\end{document}